**Non-target Structural Displacement Measurement Using Reference Frame Based Deepflow**

Jongbin Won[a], Jong-Woong Park[a*], and Do-Soo Moon[b]

[a]*School of Civil and Environmental Engineering, Chung-Ang University, Dongjak, Seoul 06974, Korea*
[b]*Department of Civil and Environmental Engineering, University of Hawaii at Manoa*
* Corresponding author: Tel: +82-2-820-5278, E-mail address: jongwoong@cau.ac.kr

*Abstract—*

Structural displacement is crucial for structural health monitoring, although it is very challenging to measure in field conditions. Most existing displacement measurement methods are costly, labor-intensive, and insufficiently accurate for measuring small dynamic displacements. Computer vision (CV) based methods incorporate optical devices with advanced image processing algorithms to accurately, cost-effectively, and remotely measure structural displacement with easy installation. However, non-target based CV methods are still limited by insufficient feature points, incorrect feature point detection, occlusion, and drift induced by tracking error accumulation. This paper presents a reference frame based Deepflow algorithm integrated with masking and signal filtering for non-target based displacement measurements. The proposed method allows the user to select points of interest for images with a low gradient for displacement tracking and directly calculate displacement without drift accumulated by measurement error. The proposed method is experimentally validated on a cantilevered beam under ambient and occluded test conditions. The accuracy of the proposed method is compared with that of a reference laser displacement sensor for validation. The significant advantage of the proposed method is its flexibility in extracting structural displacement in any region on structures that do not have distinct natural features.

## 1. Introduction

Structural health monitoring (SHM) increases a structure's lifetime and ensure its safety; the continuous monitoring provided by SHM allows for early-stage damage detection and downtime reduction, as well as potentially preventing failure during operation. For efficient monitoring, accurate and precise acquisition of structural response data is critical for condition assessment and decision-making that requires processed data. Structural displacement is one of the most important SHM factors when evaluating a structure's condition; traditionally, displacement is measured directly using a linear variable differential transformer (LVDT). One end of the LVDT is fixed to the structure and the other is attached to a stationary reference such as a scaffold. If the reference is fixed and stable, displacement can be measured to within a few micrometers. However, the use of LVDTs is hindered by practical difficulties in installing a reference point [1, 2]. Hence, measurement is limited to only several points on a structure. a laser Doppler vibrometer (LDV), another direct measurement method, can provide high-resolution noncontact displacement data [3, 4] but is cost-inefficient and restricted to measuring displacement in the direction of the emitted laser.

Alternatively, indirect methods using a global positioning system (GPS) have been proposed [5–9] for SHM; by their nature, such methods do not require a stationary reference and sensors can be instrumented without much effort. The accuracy of general GPS is applicable to structures with large deformation and it can be employed for long-term monitoring. However, GPS may not be sufficient for monitoring bridges which need few mm level measurement. Also, displacement can be obtained indirectly by double integration of acceleration data [1], but the measurement is limited to determining zero-mean dynamic displacement.

Recently, computer vision (CV) based structural health monitoring research has been an active research area in displacement measurement for SHM because of its cost-effectiveness, high resolution, and relatively simple instrumentation. CV based displacement measurement can be grouped into target- and non-target based CV systems. Target based systems employ custom designed target for accurate and robust tracking of displacement [2, 10–25]. For example, a target with a known geometry containing four white dots on a black background and a tracking algorithm that detects the center of the target using adoptive binary thresholding were employed for robust and real-time displacement measurement [2]. The target, designed with a high-contrast black and white pattern, allowed accurate displacement measurement but requires access to the structure for installation, which can limit a system's usefulness in the field.

Recently, non-target-based CV systems have been developed to capture a structure's existing features [25–30] as a target. However, detection of a structure's natural features can be very difficult due to a lack of contrast and background conditions. To address these issues, Kanade–Lucas–Tomasi (KLT) tracker [31, 32] is widely employed for non-target-based displacement measurement, as it detects features like bolts and edges based on the magnitude of the image gradient. Once features are detected, the KLT algorithm calculates optical flow, which is the velocity field of features across two input images—displacement is then obtained by integrating the Lucas–Kanade optical flow [33]. The performance of KLT tracker in structural displacement measurement has been validated experimentally [26, 34, 35]. For example, virtual vibration monitoring measurement was proposed to track feature points in selected regions on a structure to track multiple points simultaneously [34]. System identification of a shear building structure was conducted by measuring displacement in multiple regions using a Harris corner feature detector and KLT tracker [26]. However, two main challenges remain in implementing KLT tracker: the disappearance or incorrect tracking of feature points resulting from a low gradient in the given image and displacement drift due to optical flow integration.

To address these challenges, this paper proposes non-target based structural displacement measurement based on a combination of Deepflow [36], point of interest (POI) selection via masking, and signal filtering. Deepflow calculates pixelwise dense optical fields using Deepmatching [37], which finds dense matching features between two image frames. The proposed method calculate dense optical flows at certain frames in reference to an initial frame to directly measure drift-free displacement. In addition, masking and filtering methods are proposed to precisely filter the POIs for tracking while eliminating noisy and incorrect tracking results.

The remainder of this paper is organized as follows: Section 2 briefly reviews the optical flow and KLT tracker that are most widely used for displacement measurement. In Section 3, the proposed method—including POI selection via masking, Deepmatching, Deepflow, and signal filtering—is explained. Section 4 describes experimental validation of the proposed method conducted on a cantilever beam under occluded conditions. Finally, Section 5 presents conclusions drawn based on the experimental validation.

## 2. Background

*2.1. Optical flow*

Optical flow refers to a local displacement vector field of object motion between two consecutive frames, which occur because of movement by the object or the camera. Calculating optical flow requires two basic assumptions: 1) brightness constancy, which assumes that the pixel intensities of an object in an image do not change between consecutive frames, and 2) small motion between consecutive images. If a pixel in image frame *I* (*x, y, t*) moves by distance (*dx, dy*) in the next frame, taken after a period of time *dt*, the following equation can be applied under basic assumptions of optical flow:

$$I(x,y,t) = I(x+dx, y+dy, t+dt). \tag{1}$$

Expanding the first term of Eq. (1) using the Taylor series, the following equation can be obtained by removing higher-order terms and dividing by *dt*:

$$I(x+dx, y+dy, t+dt) = I(x,y,t) + \frac{\partial I}{\partial x}\frac{dx}{dt} + \frac{\partial I}{\partial y}\frac{dy}{dt} + \frac{\partial I}{\partial t} + H.O.T. \tag{2}$$

Combining Eqs. (1) and (2) leads to

$$\frac{\partial I}{\partial x}\frac{dx}{dt} + \frac{\partial I}{\partial y}\frac{dy}{dt} + \frac{\partial I}{\partial t} = 0 \ \ or \ \ I_x V_x + I_y V_y = -I_t, \tag{3}$$

where $V_x$ and $V_y$ are components of the velocity or optical flow of *I(x,y,t)*, and $I_x, I_y$ and $I_t$ are derivatives of the image at *(x,y,t)*. Equation (3) is called the optical flow equation.

*2.2. Lucas–Kanade method and KLT tracker*

Information from a single point in an image frame is not sufficient to accurately determine the optical flow vector. The Lucas–Kanade tracker assumes that there is a region of interest (ROI) in which all points have the same constant optical flow vector that satisfies the following equations:

$$\begin{bmatrix} \sum I_{x_i}^2 & \sum I_{x_i} I_{y_i} \\ \sum I_{x_i} I_{y_i} & \sum I_{y_i}^2 \end{bmatrix} \begin{bmatrix} V_x \\ V_y \end{bmatrix} = \begin{bmatrix} -\sum I_{x_i} I_{t_i} \\ -\sum I_{y_i} I_{t_i} \end{bmatrix}, \qquad (4)$$

where *i* is the index of the pixels in the window. Note that the first term on the left-hand side is a Hessian matrix in the first image, *I (x,y,t)*, which affects the stability of the solution to Eq.(4)—the inverse of the Hessian can become a singular matrix if the minimum eigenvalues are very small. The main idea behind the KLT tracker is to find only good features such that the inverse of the Hessian become nonsingular and reliable tracking can be performed. The KLT tracker allows for fast computation of optical flow, as only sets of good features are tracked across frames. However, due to the sparsity of the feature points, tracking accuracy heavily relies on the quality of feature points, which may change in appearance over time due to movement. Also, as the displacement is obtained by integrating optical flow **V** from two subsequent images, small errors may accumulate and result in drift.

## 3. Proposed method

*3.1. Overview*

Despite the fast computation of KLT, its application to displacement monitoring using KLT tracker suffers from long-term drift and loss of feature points due to object occlusion and movement. The proposed method employs Deepflow, which uses Deepmatching to obtain dense optical flow and defines POIs via masking, to extract displacement at ROIs. In addition, the proposed method resolves displacement drift by calculating optical flow for the incoming image frame in reference to the initial reference frame, yielding a direct displacement field without the numerical integration of optical flow that is implemented in the KLT method. An overview of the proposed structural displacement measurement method is illustrated in Figure 1. Calibration is conducted in the first stage to compensate for lens distortion. During initialization, the initial image frame is set as a reference and POIs that denote the pixel coordinates for displacement measurement are defined using a mask. Once the reference frame and POIs are set, the optical flow between the reference and subsequent input frames is computed using Deepmatching and Deepflow, tracking the movement of POIs. The measured displacements are filtered to eliminate noisy measurements and then averaged to provide high-accuracy displacement results.

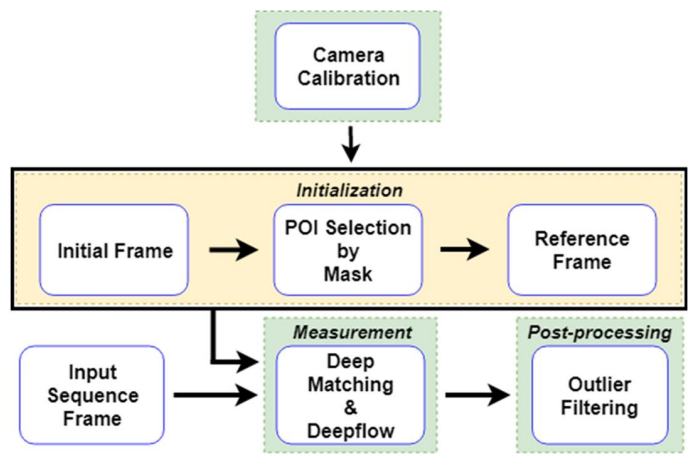

Figure 1. Flowchart of the proposed method

## 3.2. Camera calibration

Camera calibration compensates for image errors induced by lens distortion and viewing position by identifying parameters for intrinsic, extrinsic, and distortion coefficients. These parameters can be identified by employing a pinhole camera model to precisely estimate scale factor $\lambda$, intrinsic matrix $K$, translation vector **T**, and rotation matrix $R$.

$$\begin{bmatrix} x \\ y \\ 1 \end{bmatrix} = \lambda K [R|T] \begin{bmatrix} X \\ Y \\ Z \\ 1 \end{bmatrix} \tag{5}$$

The scale factor links pixels to corresponding distances in global coordinates. Intrinsic matrix $K$ is related to the camera's intrinsic properties, including focal lengths, the skew parameter, and the principal point. Extrinsic parameters are related to the physical position of the camera's view, and include the rotation matrix and translation vector. The proposed method calibrates internal parameters in the lab to compensate for distortion and obtain a scaling factor by comparing the number of pixels of a target structure in an image frame with the corresponding actual distances. The rotation matrix and translation vector are assumed to be a unit matrix and zero vector.

## 3.3. POI selection by masking

Estimating dense optical flow using correspondence in high-resolution input image or video is computationally expensive. Cropping the input image and selecting POIs as preprocessing can provide greater efficiency. The flow is estimated from the POI features within the cropped image. In the KLT method, POIs should be larger than the structural features to extract more feature points for reliable tracking; thus, it is possible to detect points outside structural areas. The proposed POI selection by masking efficiently extracts points for tracking in non-target-based CV applications, where detecting natural feature points or patterns can be challenging. Figure 2 illustrates the proposed masking method, which selects POIs using a binary mask; these points are tracked by the dense optical flow vector calculated using Deepflow, which is explained in Section 3.3. The main advantage of POI selection is to acquire dense POIs on the structure regardless of distinctive features, patterns, or textures, which is very challenging when using the KLT method.

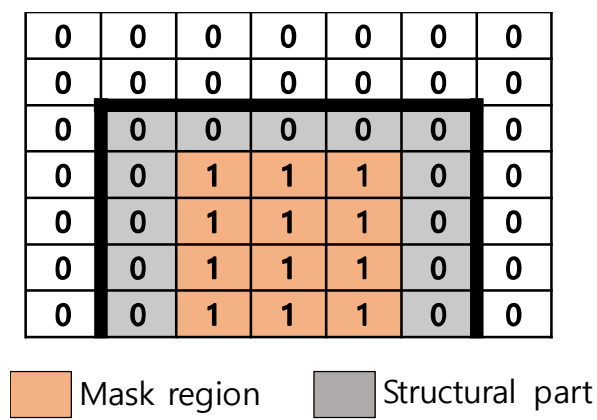

Figure 2. POI selection using masking

### 3.4. Deepmatching

Deepmatching computes dense correspondences between reference image and the target image. The matching algorithm is based on a multilayered architecture, similar to deep convolutional networks. Deepmatching splits the image at the $i^{th}$ frame into nonoverlapping 4 × 4 pixel atomic patches and convolves it with the image at the $j^{th}$ frame to obtain a response map for the corresponding image patch. This process is repeated for all patches. In the aggregation stage, response maps are max-pooled with a 3 × 3 filter and downsampled by a factor of two to reduce computational complexity. Then, average pooling is implemented for preprocessed response maps that are extracted from four neighboring patches. The final aggregation process is nonlinear filtering, which avoids fast convergence. Through aggregation, a virtual response map for 8 × 8, 16 × 16, and 32 × 32 patches is constructed; the procedure is iterated to acquire a multiscale pyramid. Note that the pyramid is built using a bottom-up approach, whereas extracting corresponding matches uses a top-down method by extracting scale-space local maxima and backtracking the configuration to obtain quasi-dense correspondences.

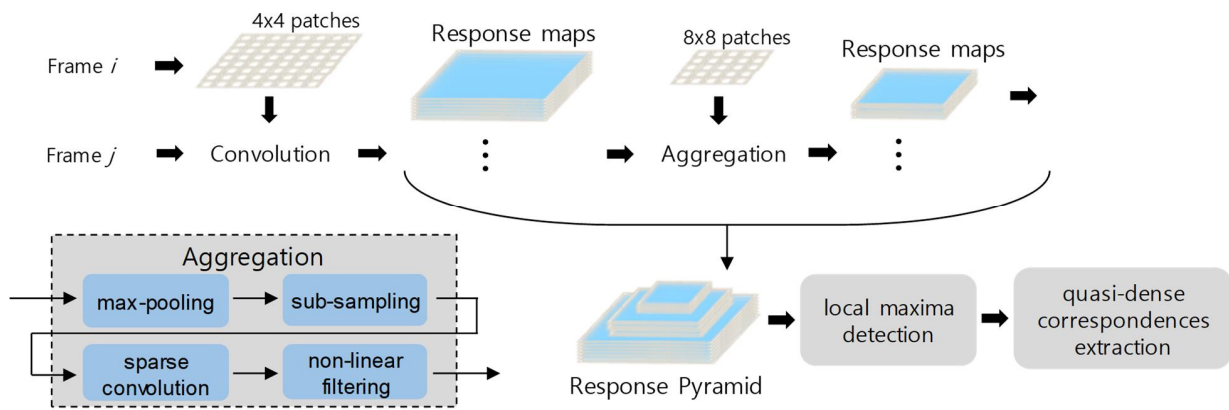

Figure 3. Deep matching flow diagram

### 3.5. Deepflow

Deepflow is a dense optical flow based on a variant method for optical flow computation that combines color and gradient constraints with a global smoothness over the computed flow field and blends the Deepmatching algorithm into an energy minimization framework. The energy to be optimized is a weighted sum of data term $E_D$, smoothness term $E_S$ and matching term $E_M$, expressed as

$$E(\mathrm{W}) = \int_\Omega E_D + \alpha E_S + \beta E_M d\mathrm{X}, \tag{6}$$

where $\mathrm{w} = (u,v)^T$ is the optical flow field, $\mathrm{x} := (x,y)^T$ denotes a point in the image domain $\Omega$, and $\alpha$, and $\beta$ are tuning parameters. Data term $E_M$ penalizes brightness and gradient constancy assumptions; it is the sum of two terms, balanced by weights $\delta$ and $\gamma$:

$$E_D = \delta\psi(|I_2(\mathrm{x}+\mathrm{w}(\mathrm{x})) - I_1(\mathrm{x})|^2) + \gamma\psi(|\nabla I_2(\mathrm{x}+\mathrm{w}(\mathrm{x})) - \nabla I_1(\mathrm{x})|^2), \tag{7}$$

where $\psi$ is a robust function that handles occlusions. The smoothness term enforces regularity by penalizing the total variation of the flow field, as

$$E_s = \psi(\|\nabla u(\mathrm{x})\|^2 + \|\nabla v(\mathrm{x})\|^2). \tag{8}$$

The matching term approximates the flow estimation to a precomputed vector field by penalizing the difference between computed vector field $W$ and precomputed vector field $W'$.

$$E_M = c\psi\varphi(\|\mathrm{w}-\mathrm{w}'\|^2), \tag{9}$$

where $c$ is a binary term with a value of 1 if a match is possible and $\phi$ is a weight term that has a low value if the match is false. An incremental coarse-to-fine warping strategy is employed to solve a nonconvex and nonlinear energy functional for Deepflow.

*3.6. Reference frame based displacement measurement*

Reference frame based displacement measurement takes the initial frame as reference and calculates its optical flow with the current frame to directly obtain a displacement field, without having to integrate the optical flow from two subsequent images as KLT methods do. This approach uses two subsequent images and can be disturbed by occlusion of the camera by obstacles and by the accumulation of tracking error at each subsequent frame, which causes displacement drift. Reference frame based displacement measurement calculates the change in the displacement of the input frame associated with the reference frame. Figure 4 shows a reference frame based measurement, where $d_m$ represents the measured displacement between the $i^{th}$ and reference frame at $m$-th POI.

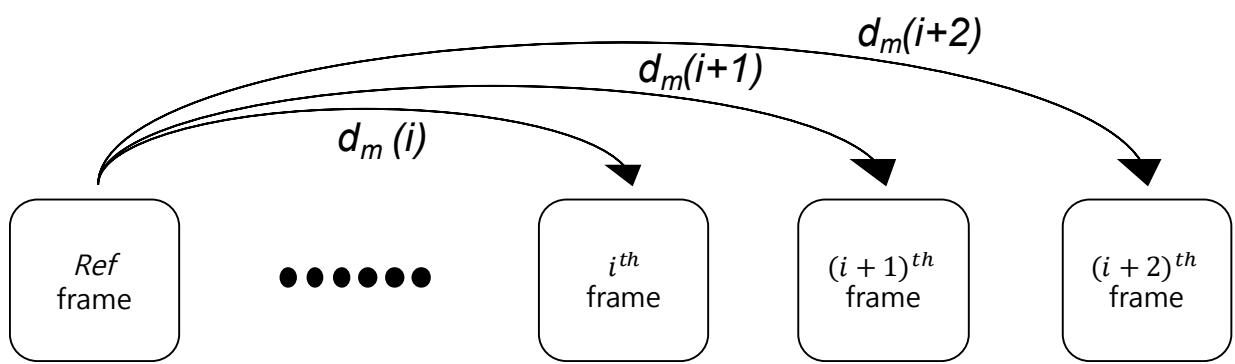

Figure 4. Reference frame based displacement measurement

*3.7. Outlier filtering and signal averaging*

Deepflow provides time-series displacement for each pixel in a selected ROI. However, the estimated flow may include outliers caused by noise, such as vanishing features or incorrectly matched feature and background points. This section proposes a filtering process for extracting accurate points related to the displacement of a target object. These points are then averaged to increase the signal-to-noise ratio of the measurement.

Let $D$ be a matrix containing offset-removed displacements at POIs, as $D=[d_1, d_2, ... d_m]$. $D$ is first threshold-filtered with its median value to remove points related to a stationary background:

$$D_{filter} = \{df_1, df_2, ... df_n\}, \begin{cases} d_m \in D_{filter} & \text{if } \max(|d_m|) \geq threshold, \\ d_m \notin D_{filter} & \text{otherwise} \end{cases} \quad (10)$$

where $Df_{ilter}$ is filtered displacements and $n$ is the number of filtered displacements. To improve measurement accuracy, outlier filtering using a correlation coefficient is adopted. The correlation coefficient matrix $R_{ij}$ between the filtered displacements is

$$R_{ij} = \mathrm{E}\left[ \left(\frac{df_i - \mu_i}{\sigma_i}\right)\left(\frac{df_j - \mu_j}{\sigma_j}\right) \right], \quad (11)$$

where $\mu_i$ and $\sigma_i$ are the mean and standard deviation of $df_i$. $MCR = \{mcr_1, mcr_2, ... mcr_m\}$, the mean of cross correlation is obtained with $m$ indicating index of points on POI. MCR over defined threshold is selected and corresponding displacements on POI are extracted and then averaged to provide a displacement with an increased signal-to-noise ratio as

$$d_{avg}(i) = \frac{1}{n_d}\sum_o df_o, \qquad (12)$$

where $d_{avg}$ is the averaged structural displacement, $n_d$ is the number of structural displacements after filtering, and $i$ is the sequence of the frame, $o$ is the number of structural displacements after filtering from $d_{filter}$.

## 4. Experiment Validation on a Steel Beam Model
### 4.1. Experimental setup

An experiment was carried out to validate the proposed non-target-based structural displacement measurement method and for comparison with KLT and laser displacement sensor methods. A subsequent experiment with environmental disturbance was implemented by blocking the camera during measurement to simulate occlusion. An overview of the experimental setup is described in Figure 5. In the experiment, a steel cantilever beam with a height of 1000 mm and a cross-section of 100 mm × 5mm was used as a testbed. The three major natural frequencies of the beam were 4.8 Hz, 24.4Hz, and 69 Hz. Video of the beam's motion was taken with a Samsung Galaxy S9+ mobile phone camera at 1 m from the beam using the 4K UHD (60 fps, 3840 × 2160 pixel resolution) setting. The reference displacement was measured using an ILD-1420 with a 1 kHz sampling rate. A logo was attached to the back of the beam to artificially introduce noise in feature tracking—only this region was cropped for efficient image processing. The scale factor (1 mm / 2.8 pixels) was obtained by comparing the beam thickness of 5 mm and the corresponding image pixels.

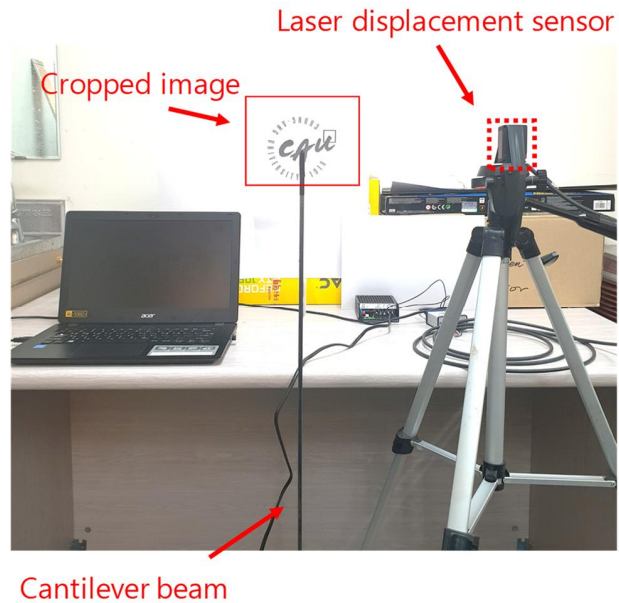

Figure 5. Experimental setup

*4.2. Feature extraction*

Figure 6 shows the cropped ROI from an image. From the image, the feature points that are tracked for displacement measurement over frames should be carefully selected for reliable tracking. The proposed masking-based POI allows selection of any points for tracking, so 77 points inside the structure were chosen for displacement tracking (see Figure 6). In the KLT method, feature points were selected based on feature detection algorithms such as Harris corner and scale-invariant feature transform (SIFT) methods using the gradient of the given image. Compared with the proposed method, the KLT method with Harris corner feature detection only detected ten points that included the edge of the beam and background features, but no features were detected inside the structure because feature point detection is heavily affected by gradient magnitude.

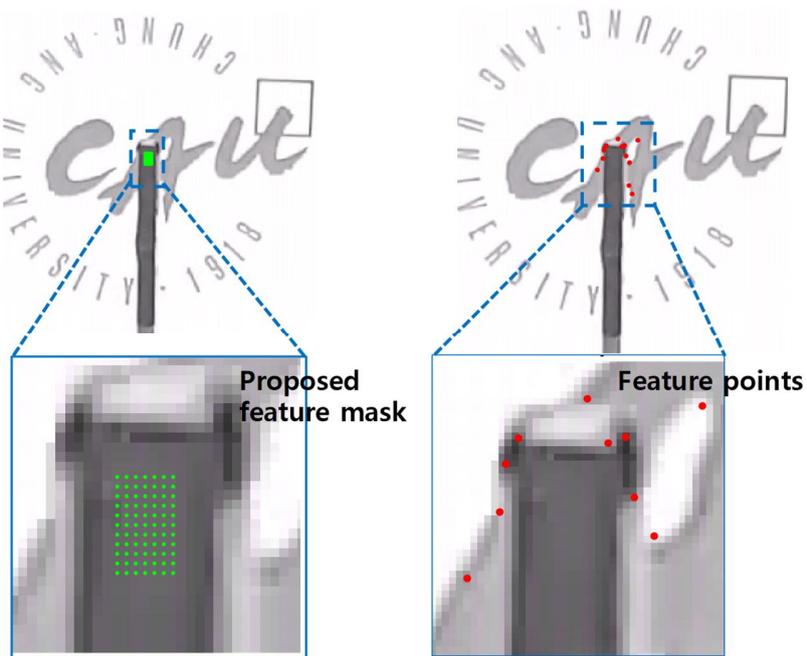

Figure 6. Feature detection in the cropped images: proposed mask (left) and Harris corner (right)

Figure 7 shows that the region inside the structure has a very small gradient magnitude, resulting in no feature detection, whereas edges and backgrounds with strong gradients match points where features were extracted.

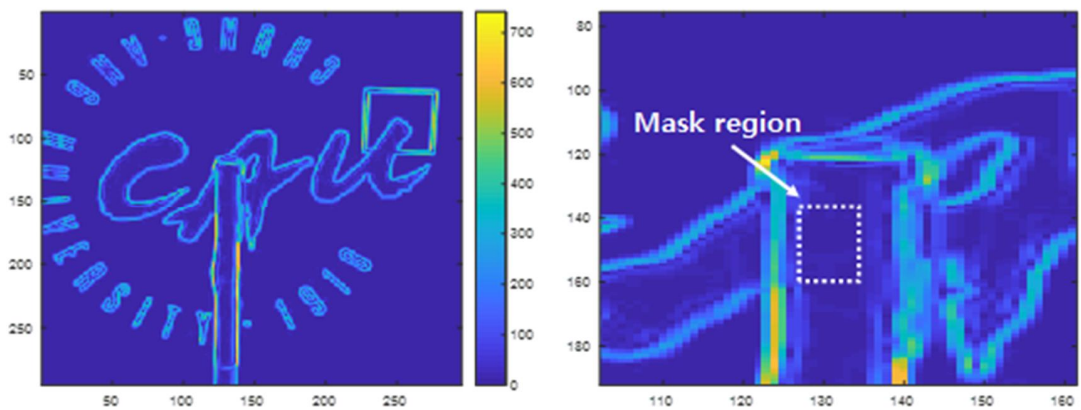

Figure 7. Gradient magnitudes in the cropped image

*4.3. Displacement measurement under ambient condition*

The KLT and proposed methods were utilized to measure the displacement of the cantilever beam model and compared with reference data measured by a laser displacement sensor. The measured displacements using the KLT and proposed methods are shown in Figure 8. To compare the result of the reference displacement sensor with those of the proposed and KLT method, a third-order Butterworth low-pass filter with a cutoff at 30 Hz was applied to the reference displacement sensor and the measurements were synchronized. The maximum and root-mean-squared errors (RMSE) of the displacements are compared in Table 1.

Table1. Comparison of the displacement measurements

| Method | Maximum displacement (mm) | RMSE (mm) |
|---|---|---|
| Proposed method | 1.7909 | 0.0753 |
| KLT | 1.5017 | 0.2943 |
| Reference displacement sensor | 1.7986 | - |

(a) 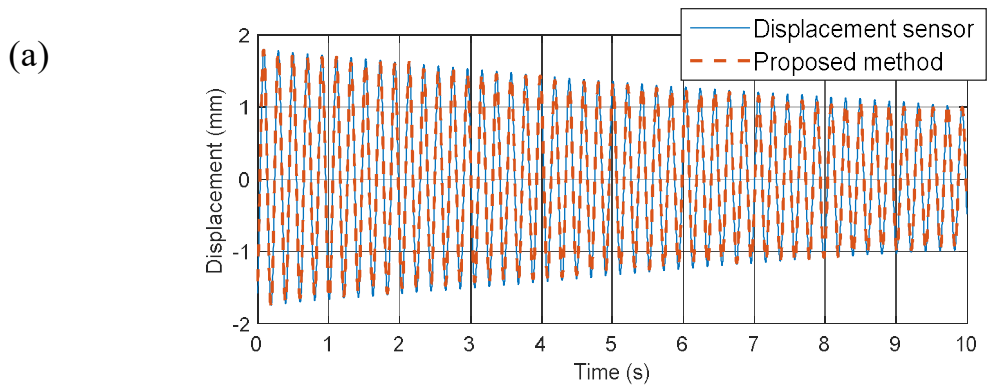

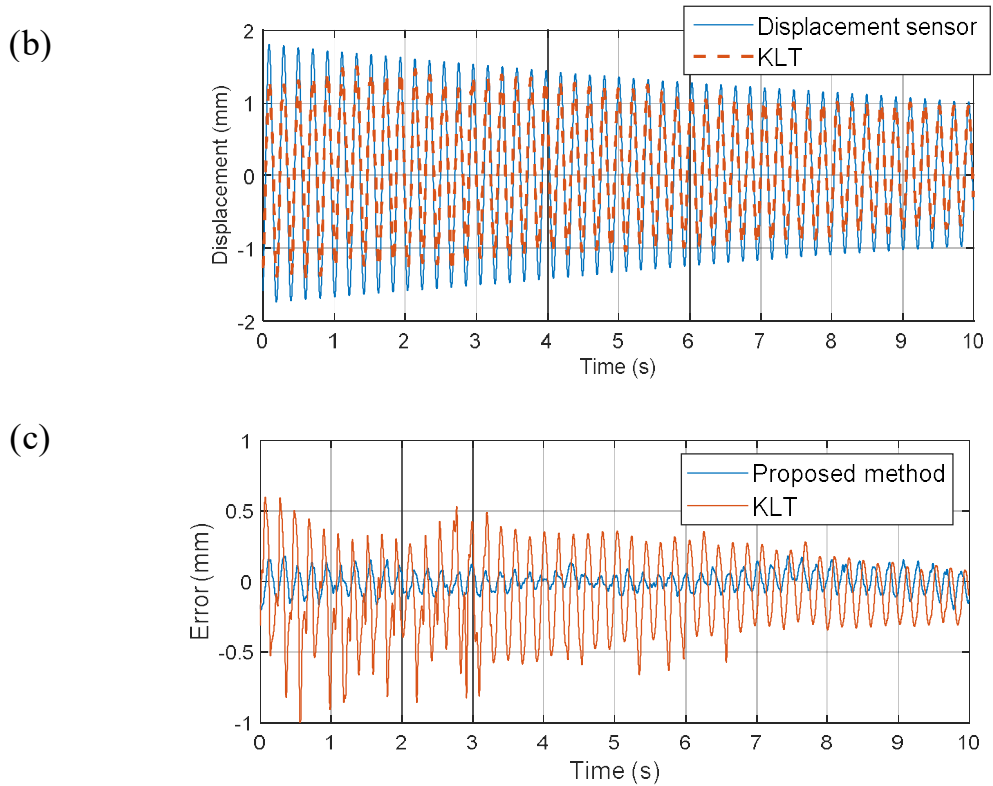

Figure 8. Optical displacement comparison: (a) displacement sensor vs. the proposed method; (b) displacement sensor vs. KLT; and (c) displacement error from (a), (b).

The proposed method showed a maximum displacement error of 0.43% compared to the reference displacement sensor, whereas the KLT method showed an error of 19.77%. Comparing RMSEs, the proposed method had a very small error (0.07 mm) validating its accuracy in measuring displacements of less than 0.1 mm, but the KLT method showed an error of 0.29 mm. The ratio of the RMSE to the maximum reference displacement was 3.80% for the proposed method and 16% for the KLT method. The KLT method showed relatively lower accuracy because of scaling errors and drift. Because the KLT detects feature points that are determined by feature detection algorithms such as Harris corner and SIFT, detected features are likely to contain background features that do not have structural motion or noisy features that are strongly affected by changes in brightness. The displacements from detected features are simply averaged without filtering, so resulting displacements become smaller than the desired structural displacement due to inclusion of the motionless background. Moreover, detection of noisy features leads to displacement drift as errors accumulate through numerical integration. The scaling error and drift are clearly identified by comparing the frequency domains in Figure 9. The magnitude of the power spectral density (PSD) at the first natural frequency at 4.8 Hz from the proposed method agrees very well with the reference displacement sensor, indicating that the dynamic responses captured and successfully identified the frequency peak. However, the magnitude of the PSD at the first natural frequency from the KLT method was smaller than that from the reference displacement, indicating a smaller displacement measurement. Furthermore, the proposed method had almost the same PSD magnitude in the 0–0.1 Hz region as the reference displacement sensor, whereas the KLT method showed a magnitude as high as the first natural frequency, indicating significant measurement drift in the resulting displacement compared to the reference displacement sensor.

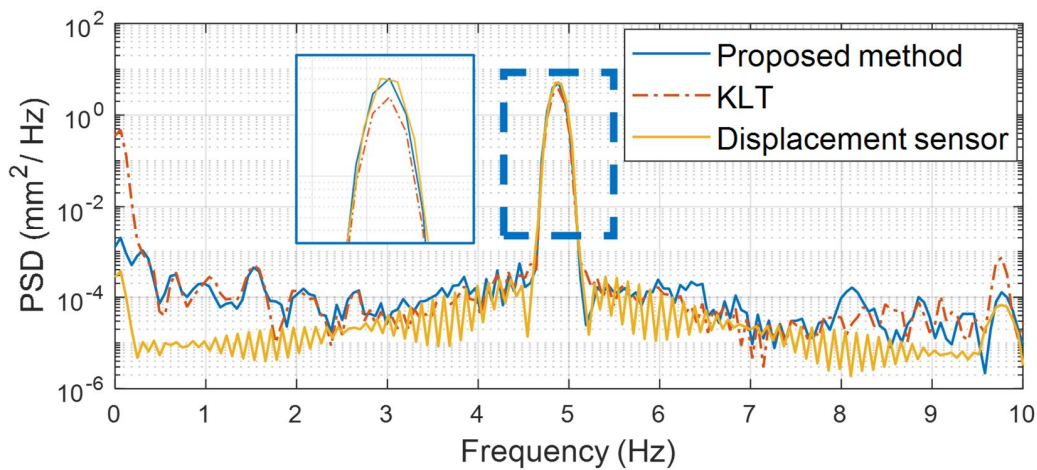

Figure 9. Frequency domain comparison using PSD

*4.4. Displacement measurement under disturbed condition*

A second experiment was conducted to determine the robustness of the proposed method to occlusion, for long-term measurements. In the field, vision systems are interfered with by many factors that block the camera's sight, which causes significant measurement errors. To implement occlusion, the camera's view was blocked with A4 white paper for about 1 s and then removed. Figure 10 shows displacement measured by the KLT and proposed methods. In figure 10(b), the KLT method measured very large displacement when occlusion occurred. Since the KLT method tracks feature points that are detected based on the difference between the structure and the surrounding background, errors were caused by mistakenly recognizing some parts of the obstacle as feature points. Additionally, after the camera's view has recovered, the feature points cannot be restored properly, resulting in an offset error. In contrast, figure 10(a) shows that the proposed method, which captures features inside the structure using a masking technique, continuously measured displacement by correctly recovering the feature points.

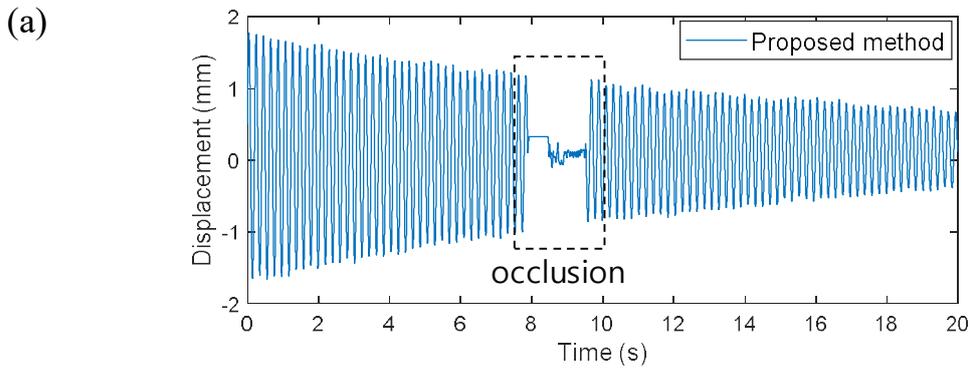

(a)

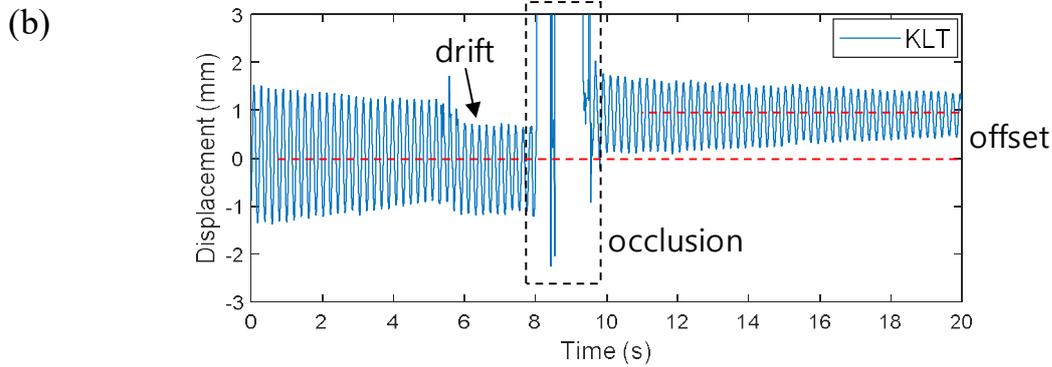

Figure 10. Comparison of optical displacement: (a) Proposed method; (b) KLT;

## 5. Conclusions

This paper proposes a non-target- and CV based structural displacement measurement system using reference frame based Deepflow, POI selection with masking, and signal filtering and averaging techniques. The proposed method directly measures displacement by calculating optical flow with a reference frame, which is updated to provide a robust tracking result. In addition, as Deepflow allows for pixelwise optical flow calculation, feature points related to structural displacement can be abundantly populated. These feature points are filtered and averaged for accurate displacement measurements while removing background noise. The proposed method was experimentally validated with a cantilever beam and its displacement result was compared with that of a laser displacement sensor. First, the proposed method was compared with KLT in stable conditions; due to some incorrect matches by the KLT method, the proposed method showed a better RMSE with 0.07 mm and 0.29 mm for proposed method and KLT respectively. Note that the KLT method showed drift over the measurement period because of erroneous feature point detection between the structure and the background. Second, displacement was measured under occluded conditions where the camera was entirely blocked for about 1 sec. During blocking, the proposed method tracked drift-free displacement under such abrupt disturbance whereas the KLT method missed or incorrectly detected feature points, resulting in significant drift and offset measurement errors. In conclusion, the ability to measure non-target-specific drift-free displacement was the most significant advantage of the proposed method, which was implemented with Deepflow, masking, and signal filtering and averaging techniques. Future work based on this study will include long-term field experiments and multiple-point tracking for system identification.

**Funding**

This work was supported by the Chung-Ang University Research Grants in 2018.